\documentclass[conference]{IEEEtran}
\IEEEoverridecommandlockouts
\usepackage{cite}
\usepackage{amsmath,amssymb,amsfonts}
\usepackage{algorithmic}
\usepackage{algorithm}
\usepackage{algorithmic}
\usepackage{graphicx}
\usepackage{graphicx}
\usepackage{textcomp}
\usepackage{xcolor}
\def\BibTeX{{\rm B\kern-.05em{\sc i\kern-.025em b}\kern-.08em
    T\kern-.1667em\lower.7ex\hbox{E}\kern-.125emX}}
\begin{document}

\title{Cross-Domain Few-Shot Relation Extraction via Representation Learning and Domain Adaptation\\
}

\author{Zhongju Yuan\textsuperscript{1}, Zhenkun Wang\textsuperscript{1,2 *}\thanks{* Corresponding author: wangzhenkun90@gmail.com.}, Genghui Li\textsuperscript{1}\\
  \textsuperscript{1}School of System Design and Intelligent Manufacturing\\
  \textsuperscript{2}Department of Computer Science and Engineering\\
  \texttt{Zhongju.Yuan@UGent.be},
\texttt{wangzhenkun90@gmail.com}\\
\texttt{genghuili2-c@my.cityu.edu.hk}\\
}


\maketitle

\begin{abstract}
Few-shot relation extraction aims to recognize novel relations with few labeled sentences in each relation. Previous metric-based few-shot relation extraction algorithms identify relationships by comparing the prototypes generated by the few labeled sentences embedding with the embeddings of the query sentences using a trained metric function. However, as these domains always have considerable differences from those in the training dataset, the generalization ability of these approaches on unseen relations in many domains is limited. Since the prototype is necessary for obtaining relationships between entities in the latent space, we suggest learning more interpretable and efficient prototypes from prior knowledge and the intrinsic semantics of relations to extract new relations in various domains more effectively. By exploring the relationships between relations using prior information, we effectively improve the prototype representation of relations. By using contrastive learning to make the classification margins between sentence embedding more distinct, the prototype's geometric interpretability is enhanced. Additionally, utilizing a transfer learning approach for the cross-domain problem allows the generation process of the prototype to account for the gap between other domains, making the prototype more robust and enabling the better extraction of associations across multiple domains. The experiment results on the benchmark FewRel dataset demonstrate the advantages of the suggested method over some state-of-the-art approaches.
\end{abstract}


\section{Introduction}
Relation extraction aims to automatically identify the relations between entities in sentences, which plays a vital role in machine reading comprehension. Relation extraction is often regarded as a multi-classification task and solved by supervised learning methods~\cite{kate2010joint, riedel2010modeling}. Especially, deep learning methods have achieved impressive performance on this kind of task. For example, Zeng et al.~\cite{zeng2014relation} first apply the deep Convolutional Neural Network (CNN) to relation extraction and obtain better performance than traditional approaches. Zeng et al.~\cite{zeng2015distant} propose the piecewise CNNs to replace the complicated preprocessing to learn features for relation extraction. Moreover, the finetuning-based representation model BERT proposed in~\cite{devlin2018bert} shows state-of-the-art performance on many classification tasks. However, these methods work based on a large amount of labeled data. When the labeled data is insufficient, their performance degenerates significantly.

Relation extraction is a core issue in many scientific fields (e.g., biomedicine and materials). However, the labeled data in such fields is always insufficient since it is expensive to obtain. Few-shot learning methods can deal with this issue since they can identify the sentences of a novel class by exploiting the already trained model (called few-shot learner) and a few labeled examples of the novel class. However, few-shot learning methods only work when the novel classes are in the same domain as those employed to train the few-shot learner~\cite{gao2019fewrel}. In other words, the few-shot learning method will fail if the novel class that needs to be identified has only a small amount of labeled data and there are no few-shot learners trained in the same domain as the novel class.

Alternatively, transfer learning methods, specifically domain adaptation methods, can also be used to deal with this issue. These methods can extract a shared feature representation of multiple different domains~\cite{ganin2016domain}. For example, Shen et al.~\cite{shen2018wasserstein} introduce Wasserstein distance to improve the generalization ability by constructing domain-invariant space between the source and target domain, where the target domain has different data but related categories with the source domain. Shi et al.~\cite{shi2018genre} employ an adversarial paradigm to extract class-agnostic features in different domains. However, these methods only work when the source and target domain classes have the same label~\cite{gao2019fewrel}.

Clearly, few-shot and transfer learning methods are inefficient for identifying a novel class with the following characteristics.
\begin{itemize}
    \item The novel class has only a few labeled samples. 
    \item The source classes with a significant amount of labeled data in the same domain as the novel classes are unavailable.
    \item The source classes that share the same label as the novel class but are from a different domain are also unavailable.
\end{itemize}

For easy description, we call the above problem the cross-domain few-shot relation extraction problem. Cross-domain few-shot learning methods have shown the potential to deal with this problem. For example, Wang et al.~\cite{wang2018adversarial} use a perceptron network as the discriminator to determine whether the domain adaptation is successful or not and an encoder BERT to extract representations from various domains. The domain adaptation is successful if the discriminator can not distinguish between these domains. Although the domain adaptation method in~\cite{wang2018adversarial} can merge data from different domains in the shared latent space learned by the encoder, it has two limitations: 1) it does not explicitly keep the geometrical structure of the classes in the source domain; 2) it does not explicitly minimize the distance between the source domain and target domain. Based on this consideration, this paper proposes a cross-domain few-shot learning method with an improved domain adaptation method to deal with the cross-domain few-shot relation extract problem.

The contributions of this paper can be summarized as follows:
\begin{itemize}
  \item Based on a few labeled samples in the source domain and a few labeled samples in the target domain, an encoder is trained to extract the relation of the unlabeled sentence in the target domain.
  \item A representation loss is proposed to make the encoder not only extract the relation of the sentence in the source domain but also keep the geometric structure of classes in the source domain. Additionally, the source domain and target domain are merged via an adversarial loss.
  \item The experimental results on the Pubmed domain and the Semeval domain show that our proposed method can significantly outperform some state-of-the-art methods on the cross-domain few-shot relation extraction problem.
\end{itemize}

\section{Related work}

In the following, the related few-shot learning methods and domain adaptation methods are reviewed in detail.

\subsection{Few-shot Learning}
\label{sec:few-shot}

Generally, the few-show learning methods can be divided into three categories~\cite{munkhdalai2017meta}: (1) data-based methods, (2) algorithm-based methods, and (3) metric-based methods.

Data-based methods augment the data with prior knowledge to overcome the difficulty of insufficient data~\cite{wu2018exploit, gao2018low, cong2020inductive}. For example, Cong et al.~\cite{cong2020inductive} assign pseudo-labels to unlabeled samples for training. It works on cross-domain classification tasks when BERT aligns the features extracted from the source sentence and the target sentence. However, it is time-consuming and requires extra space to train the model.

Algorithm-based methods use prior knowledge to search for an effective initial solution for multiple tasks simultaneously, which makes it easy to adapt to new tasks~\cite{finn2017model, yoo2018efficient}. For example, the model trained by MAML~\cite{finn2017model} can work well on new tasks after fine-tuning.
Although these methods perform well on many tasks, they cannot work well on the cross-domain relation extraction tasks~\cite{gao2019fewrel}, as they fail to reduce the discrepancy of different domains.

Metric-based methods learn an encoder based on a metric to refine the sentence embedding in the latent space such that the learned latent space can generalize to novel relations with few labeled samples in the same domain~\cite{vinyals2016matching, snell2017prototypical, triantafillou2017few, soares2019matching}. For example, the prototype network~\cite{snell2017prototypical} and the matching net~\cite{wang2018low} use Euclidean distance between sentence embedding and relation prototype to identify the relation of the sentence. Generally, these metric-based methods extract the relation of the sentence based on the prototype of the relations, and the prototype is determined by the embedding of labeled sentences in the corresponding relation. The sentences are embedded by a learned encoder. However, the learned encoder in these methods does not explicitly keep the geometric structure of the classes in the latent space. Moreover, they also can not merge different domains with significant discrepancies. Therefore, these methods usually have a good performance on relation extraction tasks with insufficient labeled data only when the tasks belong to the same domain. 

\subsection{Transfer Learning}
\label{sec:domain}

Domain adaption is a vital part of transfer learning methods, which studies how to benefit from different but related domains, and it is employed to deal with various tasks in computer vision~ \cite{yang2021free, zhao2020multi} and natural language processing~\cite{shen2018wasserstein, glorot2011domain, nguyen2014employing}. Unfortunately, some existing domain adaptation methods~\cite{nguyen2015semantic} do not be suitable for our scenario since they require a large number of labeled samples in the target domain in the training process. Although other methods do not require labeled data of the target domain in the training stage, they require different domains to have the same labels, such as comments on laptops and restaurants~\cite{fu2017domain, shen2018wasserstein, goodfellow2014generative, shi2018genre, li2018extracting}. Therefore, these methods perform well for relation extraction in the target domain only if the target and source domains are highly related. In other words, existing domain adaptation cannot obtain good results for relation extraction if there are non-overlapping relations in the target domain and source domain.

\section{Methods}
\label{methods}

Our key purpose is to improve the generalization ability of few-shot relation extraction models to arbitrary unseen domains by improving the representation of prototypes. There are two domains in the cross-domain few-shot relation extraction problem: the source domain and the target domain. 
We assume that 1) the source domain and the target domain are significantly different; 
2) the labels (relations) on the source domain and target domain are different; 
3) there are only a few labeled samples in the target domain.
To address the problem, prior knowledge is utilized to explore the connection between different relations in the source domain. And contrastive learning method is also employed to improve the geometric interpretability of the generated prototype. To bridge the gap between these domains, Wasserstein distance is used to modify the representation of prototypes.

\begin{figure*}[ht]
  \centering
  \includegraphics[scale=0.6]{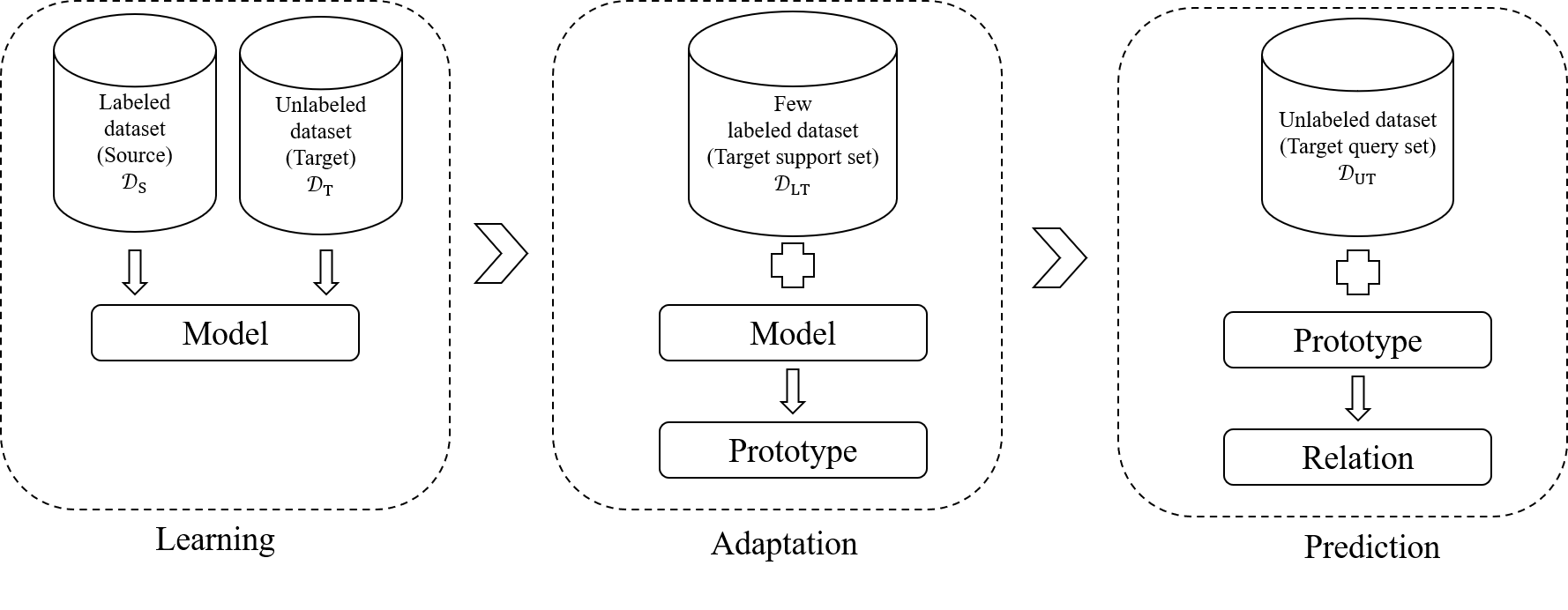}
  \caption{The structure of the proposed method}
  \label{fig:Whole}
\end{figure*}

The structure of the proposed method is illustrated in Fig.\ref{fig:Whole}, which mainly includes three phases, namely, the learning phase, the adaptation phase, and the prediction phase. In the following, we introduce them one by one. 

\subsection{Learning phase}
The learning phase is to learn an encoder to map the input sentence into the latent space. This paper adopts BERT~\cite{devlin2018bert} as the encoder. All available data of the source domain and the target domain is used to train the encoder $Enc(\cdot)$.  $\mathcal{D}_{\rm S}$ and $\mathcal{R}_{\rm S}$ denotes the sentence set and corresponding relation set of the source domain. $\mathcal{R}_{\rm S}$ includes all different relations in the source domain.     
$\mathcal{D}_{\rm T}=\{\mathcal{D}_{\rm LT},\mathcal{D}_{\rm UT}\}$, including the labeled sentence set $\mathcal{D}_{\rm LT}$ and unlabeled sentence set $\mathcal{D}_{\rm UT}$, is the sample set of the target domain. The corresponding relation set of $\mathcal{D}_{\rm LT}$ is denoted by $\mathcal{R}_{\rm LT}$. 

In order to allow the encoder to extract more interpretable prototypes that can be used to improve the relational extraction accuracy and generalizability, this paper proposes to use two loss functions $\mathcal{L}(\theta_{\mathrm{E}})$ and $\mathcal{L}_{\rm adv}(\theta_{\mathrm{E}})$ for this purpose. The representation loss $\mathcal{L}(\theta_{\mathrm{E}})$ is to make the encoder not only extract the relation of the source domain with prior knowledge but also improve the geometric interpretability of the prototypes in the source domain. And the adversary loss $\mathcal{L}_{\rm adv}(\theta_{\mathrm{E}})$ is to modify the representation of prototypes while taking the domain discrepancy into consideration. 

The representation loss $\mathcal{L}(\theta_{\mathrm{E}})$ is defined as follows.
\begin{equation}	
\mathcal{L}(\theta_{\mathrm{E}})=\mathcal{L}_{\rm cls} + \rho  \mathcal{L} _{\rm con} ,
\label{con:3}
\end{equation}
where $\mathcal{L}_{\rm cls}$ is the cross-entropy loss, $\mathcal{L} _{\rm con}$ is the proposed contrastive loss, and $\rho$ is a hyperparameter, and it is set to 0.6 based on some preliminary experiments.

Like the commonly used few-shot learning methods~\cite{gao2019fewrel}, the support set $\mathcal{S}$ and query set $\mathcal{Q}$ are randomly selected from the source domain dataset to train the encoder in each training iteration. The support set $\mathcal{S}$ includes ${\mathrm{N}}$ relations and each relation includes ${\rm K}$ sentences. The relation set of the support set is denoted as $\mathcal{R}_{\mathcal{S}}=\{r_{s}| s\in \mathcal{S}\}$. The query set $\mathcal{Q}$ includes the same $\mathrm{N}$ relations as the support set, and each relation includes $\mathrm{Q}$ sentences.

The prototype $\mathbf{v}_{r_{i}}$ plays a vital role to extract relation $r_{i}$. $\mathbf{v}_{r_{i}}, i=1,\ldots,|\mathcal{R}|$ is initialized as follows.
\begin{equation}
	\mathbf{v}_{r_{i}} = \mathbf{m}_{r_{i}} + \mathbf{h}_{r_{i}} - \mathbf{m}, \label{con:1}
\end{equation}
where $\mathbf{m}_{r_{i}}$ is the mean of the embedding set $\{\mathbf{x}_{s}|s\in \mathcal{D}_{S}, r_{s}=r_{i}\}$; $\mathbf{h}_{r_{i}}$ is the representation of the relation $r_{i}$, which is extracted by GNN from the prior knowledge $\mathcal{G}=(\mathcal{R}, \mathcal{W})$. $\mathcal{G}=(\mathcal{R}, \mathcal{W})$ denotes the global relation graph of the source domain, where  $\mathcal{R}$ includes all different relations in the source domain, and $\mathcal{W}$ consists of the link weight between relations;  $\mathbf{m}$ is the mean of the embedding of all sentences (i.e., $\{\mathbf{x}_{s}|s\in \mathcal{D}_{S}\}$) in the 
source domain. The details of the calculation for the initial $\mathbf{v}_{r_{i}}$ can refer to~\cite{qu2019gmnn}.

In the learning phase, the encoder is learned iteratively. In each iteration, a support set $\mathcal{S}$ and a query set $\mathcal{Q}$ are randomly chosen from the source domain dataset to learn the encoder. Similarly, the prototype of the relation is also updated set by set based on the Bayesian model as follows~\cite{qu2020few}.  
\begin{equation}
	{\mathbf{v}_{\mathcal{R}'}}\leftarrow{\mathbf{ v}_{\mathcal{R}'}}+\frac{\varepsilon}{2} \nabla_{{\mathbf{ v}_{\mathcal{R}'}}} {\rm log} 	p(\mathbf{v}_{\mathcal{R}'}|\mathcal{X}_\mathcal{S},\mathcal{R}_\mathcal{S},\mathcal{G})+\sqrt{\varepsilon}\widehat{z},\label{con:2}
\end{equation}
where $\mathcal{R}'$ denotes the relations sampled for the support set $S$; $\widehat{z}$ is a random noise from the standard Gaussian distribution; $\varepsilon$ is a hyperparameter, and it is set to 0.1 based on some preliminary experiments.


Based on the chain rule, the $p(\mathcal{R}_\mathcal{S}|\mathbf{x}_\mathcal{S},\mathbf{v}_{\mathcal{R}'})$ in Eq. (\ref{con:2}) can be calculated as follows. 
\begin{equation}
	p(\mathbf{v}_{\mathcal{R}'}|\mathcal{X}_\mathcal{S},\mathcal{R}_\mathcal{S},\mathcal{G}) \propto p(\mathcal{R}_\mathcal{S}|\mathcal{X}_\mathcal{S},\mathbf{v}_{\mathcal{R}'})p(\mathbf{v}_{\mathcal{R}'}|\mathcal{G}),\label{eq:post}
\end{equation}
where the $p(\mathbf{v}_{\mathcal{R}'}|\mathcal{G})$ can be seen as the prior distribution of $\mathbf{v}_{\mathcal{R}'}$ and $p(\mathcal{R}_\mathcal{S}|\mathcal{X}_\mathcal{S},\mathbf{v}_{\mathcal{R}'})$ is the conditional probability of the relation of the sentence in the support set.

The prior distribution $p(\mathbf{v}_{\mathcal{R}'}|\mathcal{G})$ of the prototype is parameterized as follows.
\begin{equation}
	p(\mathbf{v}_{\mathcal{R}'}|\mathcal{G})=\prod_{r\in\mathcal{R}'}{p(\mathbf{v}_r|\mathbf{h}_r)},\label{eq:prior}
\end{equation}
where $\mathbf{h}_{r}$ is the prototype extracted from the global relation graph $\mathcal{G}=(\mathcal{R}, \mathcal{W})$~\cite{qu2019gmnn}.

The conditional probability of the relation of the support set $p(\mathcal{R}_\mathcal{S}|\mathcal{X}_\mathcal{S},\mathbf{v}_{\mathcal{R}'})$  is estimated as follows.
\begin{equation}
\begin{aligned}
 &p(\mathcal{R}_\mathcal{S}|\mathcal{X}_\mathcal{S},\mathbf{v}_{\mathcal{R}'})
=\prod_{s\in\mathcal{S}}{p(r_s|\mathbf{x}_s,\mathbf{v}_{\mathcal{R}'})} \\
=&\prod_{s\in\mathcal{S}}\prod_{r\in\mathcal{R'}}{\frac{\mathrm{exp}(\mathbf{x}_s\cdot \mathbf{v}_{r})}{\sum_{r'\in\mathcal{R}'}{\mathrm{exp}(\mathbf{x}_s\cdot \mathbf{v}_{r'})}}}. 
\end{aligned} \label{eq:softmax}
\end{equation}

The prior knowledge is used to modify the representation of prototypes by considering the connection between relations. To explicitly maintain the geometric structure of the relations in the source domain and increase the intrinsic semantics of relations, we introduce a contrastive loss $\mathcal{L} _{\rm con}$ to deal with this issue for getting more interpretable and robust prototypes for more accurate target domain relation extraction. The contrastive loss in Eq. (\ref{con:3}) is defined as follows. 
\begin{equation}
    \mathcal{L} _{\rm con} =  \mathcal{L} _{\rm S2S} +  \mathcal{L} _{\rm S2V}, 
\end{equation}
where the $\mathcal{L} _{\rm S2S}$ means the distance between sentence embedding and the $\mathcal{L} _{\rm S2V}$ is the distance between sentence embedding and the prototype. By using this loss, we hope the learned encoder can: 1) minimize the distance between sentences in the same class; 2) minimize the distance between the embedding of sentences and their prototypes and maximize the distance between the embedding of sentences and other prototypes.

To minimize the intraclass distance between the embedding of sentences, 
$\mathcal{L}_{\mathrm{S2S}}$ is defined as follows~\cite{soares2019matching, ding2021prototypical}.
\begin{small}
\begin{equation}
	\mathcal{L} _{\rm S2S}\! =\!
	\frac{1}{{\mathrm{N}}^{2}} \sum_{i,j}  \frac{\mathrm{exp}(\delta (\mathbf{x}_{i},\mathbf{x}_{j}))}{\sum_{j'} \mathrm{exp}(1-\delta(\mathbf{x}_{i},\mathbf{x}_{j'})d(\mathbf{x}_{i},\mathbf{x}_{j'})) } ,
\end{equation}
\end{small}
where $\mathbf{x}_{i}$ is the embedding of sentence $i \in S$, and
\begin{equation}
\delta(\mathbf{x}_{i},\mathbf{x}_{j})=\left\{
\begin{array}{rcl}
	1       &      & r_{i} = r_{j}\\
	0       &      & {\rm Otherwise}
\end{array} \right. ,\\
\end{equation}
\begin{equation}
	d(\mathbf{x}_{i}, \mathbf{x}_{j})=\frac{1}{1+\mathrm{exp}(\frac{\mathbf{x}_{i}}{\Vert \mathbf{x}_{i} \Vert} \cdot \frac{\mathbf{x}_{j}}{\Vert \mathbf{x}_{j} \Vert})},
\end{equation}
where $r_{i}$ denotes the relation of sentence $\mathbf{x}_i$ in the support set; $d(\cdot, \cdot)$~\cite{soares2019matching} denotes the distance between vectors (i.e., the similarity between different vectors). 

To minimize the distance between the embedding of sentences and their prototypes and maximize the distance between the embedding of sentences and other prototypes, $\mathcal{L}_{\mathrm{S2V}}$ is defined as follows.
\begin{equation}
\begin{aligned}
 \mathcal{L} _{\rm S2V} = \frac{1}{{\mathrm{N}}^{2}} \sum_{r \in \mathcal{R}_{\mathcal{S}}}\sum_{i=1}^{\mathrm{N}*\mathrm{K}}\mathrm{log}\hat{d}(\mathbf{v}_{r},\mathbf{x}_{i}),
\end{aligned}
\end{equation}
where 
\begin{equation}
\hat{d}(\mathbf{v}_{r},\mathbf{x}_{i})=\left\{
\begin{array}{lcl}
	d(\mathbf{v}_{r},\mathbf{x}_{i})       &      & r_{i} = r\\
	1-	d(\mathbf{v}_{r},\mathbf{x}_{i})       &      & {\rm Otherwise}
\end{array} \right. .\\
\end{equation}

We enable the encoder to extract relations in the source domain effectively by minimizing $\mathcal{L}_{\theta_{\rm E}}$. Meanwhile, the accuracy of relation extraction in the target domain is improved. However, it still has the same problem with previous few-shot learning methods. Namely, they can not perform well enough when adapting to domains with large discrepancies. To deal with this issue, an adversarial loss $\mathcal{L}_{\rm adv}$ is proposed to encourage the sentences embedding in different domains as close as possible in the shared latent space. The adversarial loss $\mathcal{L}_{\mathrm{adv}}(\theta_{\rm E})$ is defined as follows.
\begin{equation}
	\mathcal{L}_{\rm adv}={\rm Wd}(\mathcal{B}_{\rm Source}, \mathcal{\tilde B}_{\rm Target}),\label{con:eq_adv}
\end{equation}
where $\mathcal{B}_{\rm Source}= \{\mathbf{x}_{1}, \cdots, \mathbf{x}_{\rm batch\_size}\}$  and $\mathcal{\tilde B}_{\rm Target}= \{\mathbf{\tilde{x}}_{1}, \cdots, \mathbf{\tilde{x}}_{\rm batch\_size}\}$ are minibatch of the sentence embedding in the source domain and target domain, respectively. ${\rm Wd}(\cdot,\cdot)$ denotes the Wasserstein distance of two subsets, 
which is illustrated below.
\begin{equation}
	{\rm Wd}_{M, \alpha}(s, t) := min_{P\in U_{\alpha}(s, t)}<P,M>
	,\label{con:eq_Wd}
\end{equation}
where $s$ and $t$ denote the distribution of the representation of sentences in source $\mathcal{B}_{\rm Source}$ and target domain $\mathcal{\tilde B}_{\rm Target}$, respectively. $P$ is a joint distribution of source and target domain, which is in the set of $U_{\alpha}(s, t)$. $M \in \mathbb{R}^{|\mathcal{B}_{\rm Source}|} \times \mathbb{R}^{|\mathcal{\tilde B}_{\rm Target}|}$ denotes the cost from the source domain to the target domain, where each element in the matrix is computed by a distance metric $M_{ij}=|\mathbf{x}_{i}-\mathbf{\tilde{x}}_{j}|^2$.

To reasonably minimize the discrepancy between the source and target domain, the Wasserstein distance (also known as Earth Moving Distance) Eq.~(\ref{con:eq_Wd}) is used here~\cite{cuturi2013sinkhorn}. The data in both domains follow a discrete probability distribution. These distributions are regarded as quality points scattered across the latent space. 

Considering the previous methods, H-divergence is a better divergence to measure the divergence as in \cite{li2018extracting}. Compared to standard L1-divergence, H-divergence limits the hypothesis to a given class, which can be better estimated with finite samples theoretically~\cite{li2018extracting}. H-divergence estimates the target error bound by learning a classifier between the source and target domains with finite (Vapnik–Chervonenkis) VC dimensions. It motivates the Domain Adversarial Neural Network (DANN)~\cite{ganin2016domain}. In reality, the neural network usually has large VC dimensions. As a result, the bound estimated by H-divergence is loose in practice.

Compared with other methods, such as the commonly used Kullback-Leibler (KL) divergence, the Wasserstein distance takes the structure of the latent space into consideration. Thus, the Wasserstein distance is able to maintain the previous geometric structure while the KL divergence cannot obtain the same performance. The similarity of data with different distributions in the same 
latent space may not be accurately measured by KL divergence. As the KL divergence between different data distributions may be the same, which cannot take the geometric structure into consideration, the ${\rm Wd}$ distance can avoid this problem.

By using the proposed method, the advantage of using contrastive loss can be enhanced. The geometric structure of the source domain will be useful for the classification of the target domain. Therefore, the representation of the sentences will gain better properties.

Finally, based on the loss $\mathcal{L}(\theta_{\mathrm{E}})$, the parameter $\theta_{\mathrm{E}}$ of the encoder is updated by the Adam optimizer~\cite{kingma2014adam}. The pseudo-code of training the encoder is shown in Algorithm 1.

\begin{algorithm}[tb]
\label{alg:algorithm}
\textbf{Input}: Data from source domain and target domain; Global relation graph $\mathcal{G}$ of the source domain; Number of relations in the support set and query set ${\mathrm{N}}$; Number of sentence(s) in the source domain ${\rm K}$; Number of sentence(s) in the query domain ${\mathrm{Q}}$; Number of epoch $E$.\\
\textbf{Output}: The parameter of the encoder $\theta_{\mathrm{E}}$ \\
\textbf{Initialization}: 
  $\mathcal{S}=\emptyset, \mathcal{Q}=\emptyset$, the prototypes $v_{\mathcal{R}}$ initialized by Eq. (\ref{con:1}).
\begin{algorithmic}[1] 
\FOR{$epoch=1,\cdots,E$}
\STATE Randomly sample ${\mathrm{N}}$ relations $\mathbf{v}_{\mathcal{R}'}=\{r_{1},\ldots,r_{\mathrm{N}}\}$ in the source domain 
\FOR{$j =1,\ldots, {\mathrm{N}}$}
\STATE {$\mathcal{S}$ $\cup$  SampleSentences($\mathbf{x}_{i}, r_{j}$), $i = 1, 2, ..., {\rm K}$}
\STATE {$\mathcal{Q}$ $\cup$ SampleSentences($\mathbf{x}_{i}, r_{j}$), $i = 1, 2, ..., {\mathrm{Q}}$}
\ENDFOR
\STATE Update prototype $\mathbf{v}_{\mathcal{R}'}$ as Eq. (\ref{con:2}).

\STATE {Compute representation loss $\mathcal{L}(\theta_{\mathrm{E}})$ by Eq. (\ref{con:3}).}
\STATE {$\theta_{\mathrm{E}}\leftarrow {\rm Adam}(\theta_{\mathrm{E}}, \nabla\mathcal{L}(\theta_{\mathrm{E}}))$}
\STATE {Extract sentence embedding in the support set sampled in the source domain $\mathcal{B}_{\rm Source}$ and a minibatch of sentence embedding in the target domain $\mathcal{B}_{\rm Target}$.}
\STATE {Compute adversarial loss $\mathcal{L}_{\rm adv}(\theta_{\mathrm{E}})$ by Eq. (\ref{con:eq_adv}).}
\STATE {$\theta_{\mathrm{E}}\leftarrow {\rm Adam}(\theta_{\mathrm{E}}, \nabla\mathcal{L}_{\rm adv}(\theta_{\mathrm{E}})$}
\ENDFOR
\caption{Training for Cross-Domain Few-Shot Relation Extraction}
\end{algorithmic}
\end{algorithm}

\subsection{Adaptation phase}
In the adaptation phase, no additional training is performed on the encoder. Instead, the already trained encoder is used as a fixed feature extractor to extract features from the input data. In this particular phase, a few labeled samples of the target domain is used to generate the prototype of the relations in the target domain based on the learned encoder. We assume that we have a labeled support set $\hat{\mathcal{S}}$ and an unlabeled query set $\hat{\mathcal{Q}}$ in the target domain. The support set $\hat{\mathcal{S}}$  includes $\hat{\mathrm{N}}$ relations, and each relation has $\hat{\rm K}$ sentences. The query set $\hat{\mathcal{Q}}$ includes some unlabeled sentences. Clearly, the prototypes $\mathbf{v}_{\hat{r}}$ generated as follows:
\begin{equation}
    \mathbf{v}_{\hat{r}}=\frac{1}{\hat{K}}\sum_{i=1}^{K}\hat{\mathbf{x}}_{i}\mathbb{I}(i,r),
\end{equation}
where $\hat{\mathbf{x}}_{i}$ is the embedding of the sentence $i$ in the support set generated by the learned encoder, and $\mathbb{I}(i,r)$ is an indicator function, defined as
\begin{equation}
\mathbb{I}(i,r)=\left\{
\begin{array}{rcl}
	1       &      & r_{i} =\hat{r}\\
	0       &      & {\rm Otherwise}
\end{array} \right.. \\
\end{equation}

\subsection{Prediction phase}
The prediction phase is to predict the relation of the sentence of the query set $\hat{\mathcal{Q}}$ in the target domain. Based on the prototype of the relation in the query set, the relation of a sentence $q$ is determined as 
\begin{equation}
     r_{q} = {\rm argmax}_{r} \frac{\hat{\mathbf{x}}_{q} \cdot \hat{\mathbf{v}}_{r}}{\sum_{i=1}^{\mathrm{N}} \hat{\mathbf{x}}_{q} \cdot \mathbf{v}_{r_{i}}}.
\end{equation}

\section{Experiments}

In this section, we conduct experiments on one benchmark dataset to evaluate our proposed approach. We make a comprehensive analysis of our approach and compare it with state-of-the-art approaches.

\subsection{Data}

In the experimental study, the FewRel dataset \cite{gao2019fewrel} is chosen, which is a widely used benchmark for few-shot relation extraction. It contains data from four different domains, including Wikipedia, SemEval-2010 task 8, NYT, and Pubmed. For our experiment setting, we use 44,800 sentences (64 classes and 700 sentences per class) from Wikipedia as the training set and 11,200 sentences (16 classes and 700 sentences per class) from Wikipedia as the validation set. And we use 1,000 sentences (10 classes and 100 sentences per class) from Pubmed. Also, we use Semeval as the testing set to conduct another experiment. The Wikipedia data serves as the source domain, while the Pubmed and Semeval data are the target domains. There are no overlapping sentences between training, validation, and testing sets. 

Beyond that, a global knowledge graph that consists of 828 unique relations in the source domain serves as the prior knowledge. The embedding of each relation in the graph has been processed by TransE algorithm \cite{bordes2013translating}. Then the graph is constructed as a 10-nearest neighbor graph as the final global relation graph in the source domain $\mathcal{G}$. The graph only contains relations in the Wikipedia dataset (source domain), which can not be used to train the model on other datasets. 

\subsection{Experimental settings}
We use accuracy as the evaluation metric in this task. The batch size for few-shot training on 5-way-1-shot and 10-way-1-shot is 4, and on 5-way-5-shot and 10-way-5-shot is 2. And the training step is 10,000 and the learning rate is 1e-1 with SGD. We follow
the original meaning of N-way-K-shot in the paper~\cite{han2018fewrel} and for more details please refer to~\cite{qu2020few}.

\subsection{Comparison and Analysis}
We choose the following methods for comparison.

\textbf{Proto} \cite{snell2017prototypical}: The algorithm of the prototype network. A few-shot relation extraction method extracts relations by measuring the distance between the sentence embedding and the prototype.

\textbf{Proto+adv} \cite{gao2019fewrel}: The Proto algorithm uses a discriminator to adjust the source and target domains.

\textbf{MTB} \cite{soares2019matching}: The algorithm, called Matching The Blanks, builds task-agnostic relation representations solely from the entity-linked text.

\textbf{GNN} \cite{garcia2017few}: The algorithm uses Graph Neural Network (GNN) to predict the relation.

\textbf{MAML} \cite{finn2017model}: The algorithm, called Model-Agnostic Meta-Learning, solves few-shot learning problems by a meta-learning method.

\textbf{Siamese} \cite{koch2015siamese}: The algorithm uses temporal CNN and an attention mechanism for few-shot learning.

\textbf{DaFeC} \cite{cong2020inductive}: The algorithm improves domain adaptation performance for few-shot classification via clustering.

\textbf{REGRAD} \cite{qu2020few}: The algorithm completes the few-shot relation extraction task via Bayesian meta-learning on the relation graph.

\textbf{REGRAD+adv} \cite{qu2020few}: The algorithm adds an adversarial part to the REGRAD model.

\begin{table}[ht]
    \centering
	\begin{tabular}{l|llll}
		\hline
		{\textbf{Alg}} &
		\textbf{\begin{tabular}[c]{@{}l@{}}5-way\\ 1-shot\end{tabular}} &
		\textbf{\begin{tabular}[c]{@{}l@{}}5-way\\ 5-shot\end{tabular}} &
		\textbf{\begin{tabular}[c]{@{}l@{}}10-way\\ 1-shot\end{tabular}} &
		\textbf{\begin{tabular}[c]{@{}l@{}}10-way\\ 5-shot\end{tabular}} \\ \hline
		Proto      & 66.22 & 77.47 & 49.77 & 65.63 \\
		Proto+adv  & 41.09 & 67.26 & 28.32 & 40.01 \\
		MTB        & 48.89 & 66.78 & 37.23 & 51.29 \\
		GNN        & 36.44 & 37.19 & 26.00 & 28.07 \\
		Siamese    & 59.60 & 78.09 & 49.33 & 65.75 \\
		MAML       & 66.62 & 78.53 & 51.90 & 65.57 \\ 
		DaFeC      & 30.21 & 30.51 & 15.17 & 17.27 \\
		Regrad     & 73.22 & \textbf{83.12} & 63.47 & 71.59 \\
		Regrad+adv & 65.10 & 71.61 & 56.44 & 56.71 \\  \hline
		\textbf{ours}  & \textbf{73.75} & 82.24 & \textbf{66.94} & \textbf{74.53} \\ \hline
	\end{tabular}
	\caption{\textbf{Results of the cross-domain few-shot relation extraction on the Pubmed dataset.} We run all the algorithms on the same conditions.}
	\label{tab:results_1}

	\begin{tabular}{l|llll}
		\hline
		{\textbf{Alg}} &
		\textbf{\begin{tabular}[c]{@{}l@{}}5-way\\ 1-shot\end{tabular}} &
		\textbf{\begin{tabular}[c]{@{}l@{}}5-way\\ 5-shot\end{tabular}} &
		\textbf{\begin{tabular}[c]{@{}l@{}}10-way\\ 1-shot\end{tabular}} &
		\textbf{\begin{tabular}[c]{@{}l@{}}10-way\\ 5-shot\end{tabular}} \\ \hline
		Proto      & 41.39 & 59.51 & 27.62 & 42.96 \\
		Proto+adv  & 26.96 & 48.06 & 13.15 & 28.19 \\
		MTB        & 35.73 & 46.87 & 30.48 & 29.64 \\
		GNN        & 32.13 & 37.12 & 14.71 & 17.92 \\
		Siamese    & 41.67 & 53.57 & 28.06 & 39.52 \\
		MAML       & 42.75 & 52.87 & 27.89 & 43.06 \\ 
		DaFeC      & 24.72 & 25.98 & 11.17 & 13.37 \\
		Regrad     & 49.98 & 67.39 & 38.19 & 50.52 \\ 
		Regrad+adv    & 50.71 & 65.46 & 38.61 & 54.56 \\  \hline
		\textbf{ours} & \textbf{52.98} & \textbf{68.45} & \textbf{39.31} & \textbf{56.65} \\ \hline
	\end{tabular}
	\caption{\textbf{Results of the cross-domain few-shot relation extraction on the Semeval dataset.} We run all the algorithms on the same conditions.}
	\label{tab:results_2}
\end{table}

As there are few studies on the cross-domain few-shot relation extraction task, the state-of-the-art algorithms in the few-shot relation extraction task and few-shot relation extraction algorithm together with the adversarial part~\cite{gao2019fewrel, qu2020few, cong2020inductive} are chosen as the baseline in this paper. The Regrad and Regrad+adv algorithms are re-implemented as the paper~\cite{qu2020few}. The DaFeC algorithm is re-implemented as the paper~\cite{cong2020inductive}. Other algorithms are re-implemented by Gao et al.~\cite{gao2019fewrel}. Bert$_{\mbox{\scriptsize base}}$ is used as the encoder to project the sentences into the latent space for all algorithms. Besides, the hyper-parameters used in our method remain the same with the setting of~\cite{gao2019fewrel,qu2020few}.

The prediction accuracy of the target domain is used as the criterion to judge the performance of the algorithms. The comparison results are shown in Table \ref{tab:results_1} and Table \ref{tab:results_2}.

All methods were reproduced under the same conditions. Notably, the encoder was not trained during the adaptation phase. However, the performance of these baselines was not competitive for the cross-domain few-shot relation extraction task. Specifically, GNN and DaFeC showed less effectiveness in solving this task, while methods specifically designed for few-shot tasks, such as Proto and MTB, did not perform well. In comparison, our approach better generalized to different domains when compared to other meta-learning methods, such as MAML and Siamese. Although the Regrad method was the most competitive algorithm, it failed to outperform our method in most situations for the cross-domain task. Previous adversarial methods only merged source and target domains by utilizing a discriminator. Despite adversarial methods optimizing model parameters twice, our method outperformed them in terms of accuracy. However, the effectiveness of our method is highly correlated with the dataset and algorithm used. In other words, the performance of our algorithm may decrease when applied to certain datasets and algorithms.

Our model surpasses state-of-the-art models because our model can ensure a better geometric structure of the latent space. The distance between sentence embedding in the same class is closer, and the distance in different classes is farther. In addition, when the prototype is the relation representation of a given sentence embedding, the distance between the sentence embedding and the prototype is closer, otherwise, it is farther. Besides, by optimizing the adversarial loss, the distribution of the target domain is as close as possible to the source domain. Thus, the performance is further improved in the cross-domain relation extraction task. 

\subsection{Ablation Study}
In this subsection, we study the impact of contrastive loss and adversarial loss on generalization accuracy. The model only optimizes cross-entropy loss $\mathcal{L}_{\rm cls}$ is named as the original model here. We conduct some ablation study on the FewRel dataset, where we compare three variant methods, i.e., original model with $\mathcal{L}_{\rm Wd}$, with $\mathcal{L}_{\rm con}$ and with both of the loss. The results are presented in Table~\ref{tab:ab_1} and Table~\ref{tab:ab_2}.

Based on the results presented in Table~\ref{tab:ab_1} and Table~\ref{tab:ab_2}, it is evident that the original algorithm performs poorly on both datasets, especially on the 10-way few-shot task. The addition of $\mathcal{L}_{\rm Wd}$ has a mixed impact, resulting in a slight improvement on the 5-way few-shot task, but no improvement or even a decline in performance on the 10-way few-shot task. Conversely, adding $\mathcal{L}_{\rm con}$ generally enhances the performance on both datasets. Combining both $\mathcal{L}_{\rm Wd}$ and $\mathcal{L}_{\rm con}$ results in improved performance on the 5-way few-shot task for both datasets, but the impact on the 10-way few-shot task is mixed, with improved performance on one dataset and deteriorated performance on the other. Overall, adding $\mathcal{L}_{\rm con}$ leads to improved performance on both datasets, while the effectiveness of the $\mathcal{L}_{\rm Wd}$ depends on the specific dataset and few-shot task. Combining both modules can further improve performance. 

Thus, we find that contrastive loss effectively improves the performance of the target domain by utilizing the geometric structure of the latent space~\cite{ding2021prototypical}. Moreover, the adversarial loss further improves the performance of the target domain by reducing the discrepancy between the source and target domains. The observation shows that combining both of the losses can help the method solve the cross-domain few-shot relation extraction problem well.

\begin{table}[]
    \centering
	\begin{tabular}{l|llll}
		\hline
		{\textbf{Alg}} &
		\textbf{\begin{tabular}[c]{@{}l@{}}5-way\\ 1-shot\end{tabular}} &
		\textbf{\begin{tabular}[c]{@{}l@{}}5-way\\ 5-shot\end{tabular}} &
		\textbf{\begin{tabular}[c]{@{}l@{}}10-way\\ 1-shot\end{tabular}} &
		\textbf{\begin{tabular}[c]{@{}l@{}}10-way\\ 5-shot\end{tabular}} \\ \hline
		Original     & 73.22 & \textbf{83.12} & 63.47 & 71.59 \\
		With\ Wd   & 72.65 & 82.97 & 61.94 & 72.47 \\
		With\ con   & \textbf{74.11} & 80.54 & 62.75 & 73.94 \\
		With\ Wd\ and\ con   & 73.75 & 82.24 & \textbf{66.94} & \textbf{74.53} \\ \hline
	\end{tabular}
	\caption{Ablation results of the cross-domain few-shot relation extraction on the Pubmed dataset.}
	\label{tab:ab_1}

	\begin{tabular}{l|llll}
		\hline
		{\textbf{Alg}} &
		\textbf{\begin{tabular}[c]{@{}l@{}}5-way\\ 1-shot\end{tabular}} &
		\textbf{\begin{tabular}[c]{@{}l@{}}5-way\\ 5-shot\end{tabular}} &
		\textbf{\begin{tabular}[c]{@{}l@{}}10-way\\ 1-shot\end{tabular}} &
		\textbf{\begin{tabular}[c]{@{}l@{}}10-way\\ 5-shot\end{tabular}} \\ \hline
		Original     & 49.98 & 67.39 & 38.19 & 50.52 \\
		With\ Wd     & 51.46 & 67.88 & 38.46 & 54.09 \\
		With\ con    & 51.79 & 67.50 & \textbf{40.87} & 55.49 \\
		With\ Wd\ and\ con   & \textbf{52.98} & \textbf{68.45} & 39.31 & \textbf{56.65} \\ \hline
	\end{tabular}
	\caption{Ablation results of the cross-domain few-shot relation extracttion on the Semeval dataset.}
	\label{tab:ab_2}
\end{table}

\section{Conclusion}

In this paper, we have proposed a novel method by integrating the method of few-shot learning and domain adaptation to solve the cross-domain few-shot relation extraction task. To improve the interpretability of the representation of prototypes, we have designed a representation loss, including a cross-entropy loss and a contrastive loss. Besides, an adversarial loss has been further employed to consider the discrepancy between different domains. Extensive experiments have demonstrated that our method performs better than some existing state-of-the-art relation extraction methods. Moreover, the effectiveness of each used loss also has been validated by experiment.

\bibliographystyle{IEEEtran}
\bibliography{ref}

\vspace{12pt}
\color{red}

\end{document}